\documentclass[twocolumn]{article}
\usepackage[a4paper, total={6in, 8in}]{geometry}
\usepackage{NLP}
\usepackage{graphicx}
\graphicspath{ {./images/} }
\title{\textbf{Predict Emoji Combination with Retrieval Strategy}}

\author{
\begin{tabular}{c@{\ \ \ }ccc}
Weitsung Lin &
\begin{minipage}{8.3em}
  \begin{center}
    Tinghsuan Chao
  \end{center}
\end{minipage} &
\begin{minipage}{8.3em}
  \begin{center}
    Jianmin Wu
  \end{center}
\end{minipage} &
Tianhuang Su \\[4pt]
\multicolumn{4}{c}{EBG, Baidu Inc, Shenzhen, 518052} \\
\vspace{-4ex}
\end{tabular}}
\date{\texttt{\{v\_liweicong, v\_zhaotingxuan, wujianmin01, sutianhuang\}@baidu.com}}

\begin{document}
\maketitle
\begin{abstract}
\begin{quote}
As emojis are widely used in social media, people not only use an emoji to express their emotions or mention things but also extend its usage to represent complicate emotions, concepts or activities by combining multiple emojis. In this work, we study how emoji combination, a consecutive emoji sequence, is used like a new language. We propose a novel algorithm called \textbf{Retrieval Strategy} to predict what emoji combination follows given a short text as context. Our algorithm treats emoji combinations as phrase in language, ranking sets of emoji combinations like retrieving words from dictionary. We show that our algorithm largely improves the F1 score from 0.141 to 0.204 on emoji combination prediction task.
\end{quote}
\end{abstract}
\section{Introduction}
Emojis are very popular and widely used in large social media platforms like Twitter, Facebook and Instagram. These visual symbols can be used to convey emotions and underlying information and strengthened its power by combining multiple ones (Figure 1). In Natural Language Processing (NLP) perspective, they provide extra information for the semantics of sentences.\\
\begin{figure}[h]
\includegraphics[width=0.48\textwidth]{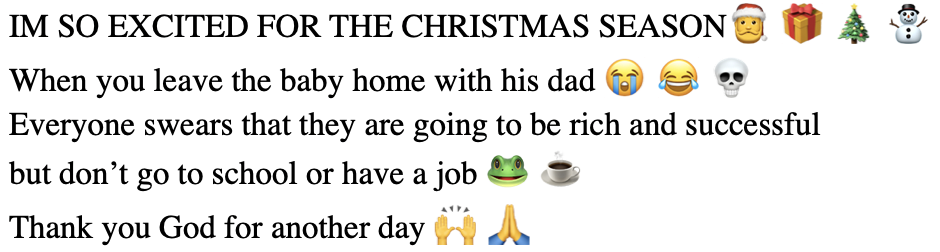}
\caption{Emoji combination use cases}
\end{figure}\\
Recent work \cite{Barbieri2017AreEP} has shown that textual information can be used to predict single emoji associated to text. \cite{Barbieri2018MultitaskEL} further extended the emoji prediction task to 300 emoji classes and used a multi-task approach, predicting emojis and emoji categories to improve prediction accuracy.
In this paper, we extend the single emoji prediction task to a setup much closer to real world usage. First, we extend emoji classes to predict to 500, which cover 95\% of emoji usage. Second, we predict multiple emojis for a given text since in real world user often use multiple emojis.\\
There is huge variation in people's usage of emoji. It could have many different but reasonable target emojis conditions on the same textual input which increases the difficulty of predicting. This problem becomes more critical in the multiple emojis task when possible candidates increase exponentially by combining multiple emojis.
To tackle this difficult problem, we propose a state of the art classification NLP model to predict the emojis from a Twitter message. This model is based on BERT \cite{Devlin2018BERTPO} with retrieval strategy we proposed.

\section{Related Works}
Previous works on emoji prediction focus on predicting single emoji by textual inputs. This is done by modeling the input text into a sentiment represent and predicting only popular emojis. The model architecture includes SVM, LSTM and LSTM with attention have been explored \cite{Barbieri2017AreEP, ltekin2018TbingenOsloAS, Baziotis2018NTUASLPAS}.\\
The recent approaches add other input modalities and predict target base on this setting. \cite{Barbieri2018MultimodalEP, Cappallo2018TheNM} consider both textual and visual inputs and build a multimodal emoji prediction. \cite{Barbieri2018MultitaskEL} uses both emojis and emoji categories from keyboard session as predicting targets to train a multi-task model and further increases the emoji labels to 300 which is closer to practical usage.\\
Our task setting is most similar to \cite{Barbieri2018MultitaskEL}, which uses only text input. The difference is in targets, we use multiple emojis  other than single emoji. This target setting is more closer to the real world use case.

\section{Dataset and Task}\label{Dataset and Task}
In this work we retrieved 10 million tweets with emojis in text from Twitter. Tweets is restricted to have geo-location in the United States of America and written in English. In the preprocessing stage, we removed all hyperlinks, mentions and hashtags from tweet. Also we normalize emoji by removing skin tone modifier to focus on meaning of emoji. To build the dataset, whose sample consists of a pair of textual context (feature) and emoji combination (target), we process Tweets as follow: (1) Select tweets include emojis in the 500 most frequent ones (2) Find consecutive sequence emojis as target and all text before it including emojis as textual context and (3) Keep only samples with targets included one to three emojis. In the end, we collect a final dataset contained about 9 million samples.\\
\begin{figure}
\includegraphics[width=0.46\textwidth]{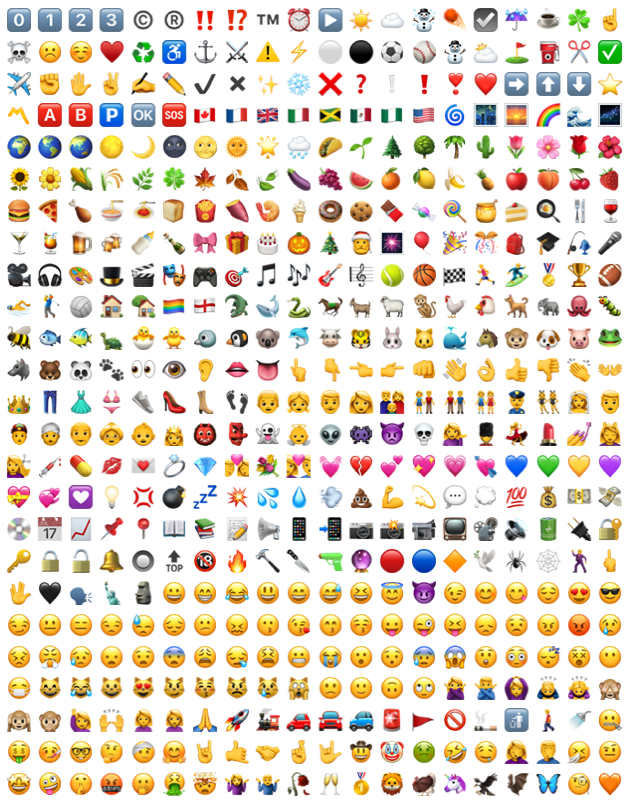}
\caption{500 emoji classes in this work}
\end{figure}\\

\section{Method}
In this Section, we describe our method to predict emoji combination given a textual context. The method consists of two parts: (1) A model to predict probability of emojis given a textual context and (2) a strategy to consume probability and generate final emoji combination.\\
\subsection{Emoji Prediction Model: BERT}
BERT is a multi-layer bidirectional Transformer encoder based on the original implementation described in \cite{Vaswani2017AttentionIA}. It obtains state-of-the-art results on NLP benchmark GLUE. We follow the training procedure described in paper with pre-training stage and fine-tuning stage.\\

\textbf{Pre-training:} In the pre-training stage, there are two differences between our model and the original BERT model.
The first one is training data. Instead of using huge corpus like Wikipedia, we only use corpus from Twitter to learn a language model because we want the language model to be more related to twitter usage and restriction of computation resource.\\
The second difference is that we only use Masked LM as pre-training target, because it is hard to determine next sentence between tweets and most tweets contain only one sentence. In model architecture, we use not only word level embedding and also character level embedding in order to capture meaning of out of vocabulary words since tweets contain lots of new words and misspellings.\\

\textbf{Fine-tuning:} In the fine-tuning stage, we select first time step vector in output sequence as input text representation like original BERT. The objective of emoji combination task is similar to language modeling to maximize following conditional probability:
\begin{equation}
    P(Y|W) = P(y_0|W)\prod_{i=1}^nP(y_i|W, y_0,..., y_{i-1})
\end{equation}
Where $W$ is the input text, $Y$ is emoji combination and $y_i$ is an individual emoji in $Y$. Because our task selects emoji combination with size between one to three and ease of computation, we approximate the n-gram conditional probability by using the following unigram models:
\begin{equation}\label{eq2}
    P(Y|W) \approx \prod_{i=0}^nP(y_i|W)
\end{equation}
We apply softmax on the model output to compute the probability distribution and minimize the cross-entropy between prediction and label distribution to maximize the log-likelihood of $\prod P(y_i|W)$.
\subsection{Retrieval Strategy}\label{Retrieval Strategy}
The naive way to produce emoji combination from probability distribution is picking top ones and finding a good threshold to produce emoji combination with variable length. However, this strategy tends to predict common emojis and produce unreasonable combinations, resulting bad performance. Moreover, the ordering of emoji combination can't be carefully modeled, hence visual meaning lost. To tackle this problem, we treat emoji as a language: taking emoji as word then emoji combination naturally becomes phrase. Instead of predicting emoji combination directly, we use retrieve strategy to rank the existed emoji combinations which are mined from corpus. In other word, we predict the emoji combination from a collection like predicting phrase from a dictionary.\\
The ranking score we used is cross-entropy between the emoji combination candidate and the prediction probability distribution:
\begin{equation}\label{eq3}
    S_j = - \sum_i Candidate_{j}(y_i) \log P(y_i|W)
\end{equation}
Where $S_j$ is the ranking score, $Candidate_{j}(y_i)$ is emoji distribution of $y_i$ in candidate $j$, $P(y_i|W)$ is the model prediction for emoji distribution. Because of the approximation we used in Eq.\ref{eq2}, emoji combination with the same emoji set but different orderings would have the same ranking score. We use frequency of emoji combination in training data to break a tie. As a result, our method can produce meaningful emoji combination contains rare emojis.
\section{Experiment}
To evaluate the quality of emoji combination prediction, we randomly select 10000 samples from the Twitter dataset for testing and use precision, recall and F1 score as evaluation metrics. We compared the emoji combination prediction task with three combination strategies: \textbf{Naive Top-3}: Select first three emojis with top probability, \textbf{Greedy Top-3}: Select emojis with top probability greedily until cumulative probability reaches threshold or size exceeds three and \textbf{Retrieval Strategy} described in section \ref{Retrieval Strategy}. In \textbf{Retrieval Strategy}, we use Eq.\ref{eq3} to rank 30k most frequently used emojis in dataset and add an emoji size penalty to trade off between recall and precision.\\
For data preprocessing, we simply use whitespace as separator for processing input text and treat each emoji in input text as a word token. The emoji is normalized by cleaning the different skin tones and only used which is in \footnote{https://unicode.org/emoji/charts/full-emoji-list.html}Unicode Full Emoji List v11.0.
In this work, the emoji prediction model has smaller architecture than BERT base model to save computation power. We denote Transformer blocks as L, hidden size as H and self-attention heads as A. The model size is: L=10, H=256, A=4, feed-forward/filters=3H and the maximum sequences length is 24 tokens.\\
In the pre-training stage, the training sequence is generated from all 10M tweets mention in Section \ref{Dataset and Task}. We train with batch size of 64 sequences for 1,500,000 steps, which is approximately 10 epochs over the 10M tweets data. We set hyperparameters, including Adam, L2 weight decay, linear decay of learning rate, dropout and activation function the same as the original BERT model.\\
In the fine-tuning stage, we take the first time step of transformer output sequence and denote this vector as $C \in R^H$. In order to calculate label probability, we add a fully-connection layer with parameters $W \in R^{K\times H}$ and $b \in R^K$, where $K$ is the number type of emojis which in our case is 500. The label probabilities $P \in R^K$ are computed with a standard softmax, $P = softmax(CW^T + b)$. The training setting is batch size 64, Adam learning rate 5e-5 and train for 5 epochs. We also add a weighted object Mask LM from pre-trained as auxiliary object like OpenAI GPT \cite{Radford2018ImprovingLU}.\\
The comparison result of combination strategies is listed in Table \ref{table1}. All strategies use the same emoji prediction model weights and only different in the combination strategy which processes emoji probability from the same model.\\
We add a value for compromising between recall and precision for both strategies. The highest F1 score is between the most balance recall and precision. The result shows our retrieval algorithm largely improved the F1 score in emoji combination prediction task.
\begin{table}[t]
\begin{center}
\caption{Combination strategies evaluation results on test dataset. The higher threshold for Greedy Top-3 results in higher recall but lower precision. The penalty in Retrieval strategy penalizes less emoji number in combination. Fewer emoji in combination results bigger penalizing. The highest F1 is Retrieval Strategy with penalty at 0.3.}
\label{table1}
\begin{tabular}{c|ccc}
 \hline
  Combination & recall & precision & F1 \\
  Strategy & (\%) & (\%) & ($\times$100) \\
 \hline
  Naive Top-3 & 29.4 & 9.3 & 14.1 \\
  Greedy Top-3 thr=0.4 & 28.0 & 13.4 & 18.1 \\
  Greedy Top-3 thr=0.3 & 25.6 & 14.5 & \underline{18.5} \\
  Greedy Top-3 thr=0.2 & 21.4 & 16.0 & 18.3 \\
 \hline
  Retrieval & 18.6 & 22.5 & 20.4 \\
  Retrieval pen=0.2 & 24.3 & 21.9 & 23.0 \\
  Retrieval pen=0.3 & 26.6 & 21.2 & \underline{\textbf{23.6}} \\
  Retrieval pen=0.4 & 28.3 & 20.1 & 23.5 \\
 \hline
\end{tabular}
\end{center}
\end{table}
\section{Conclusion}
As emojis getting widely used in social media, people are continuing inventing new emoji combinations to express different meanings. The emoji combinations are used more and more like a new language, constructed in different ordered even with structure to express new meanings. To tackle the emoji combination prediction problem, we propose a novel algorithm consists of a state-of-the-art language modeling to predict emoji probability distribution and a retrieval strategy which approximates complex n-gram model with unigram fashion to produce reasonable emoji combination. We showed that our method beats other methods on emoji combination prediction task with large margin. In addition, we add a penalty in emoji combination scoring by compromising between recall and precision for the retrieval strategy to further improve F1 score. To our best knowledge, this is the first work to study emoji combination and achieve decent result on Twitter dataset.

\bibliographystyle{plain}
\bibliography{ref}

\begin{thebibliography}{1}

\bibitem{Barbieri2018MultimodalEP}
Francesco Barbieri, Miguel Ballesteros, Francesco Ronzano, and Horacio Saggion.
\newblock Multimodal emoji prediction.
\newblock In {\em NAACL-HLT}, 2018.

\bibitem{Barbieri2017AreEP}
Francesco Barbieri, Miguel Ballesteros, and Horacio Saggion.
\newblock Are emojis predictable?
\newblock In {\em EACL}, 2017.

\bibitem{Barbieri2018MultitaskEL}
Francesco Barbieri, L{\'u}ıs Marujo, Pradeep Karuturi, and William Brendel.
\newblock Multi-task emoji learning.
\newblock 2018.

\bibitem{Baziotis2018NTUASLPAS}
Christos Baziotis, Nikos Athanasiou, Georgios Paraskevopoulos, Nikolaos
  Ellinas, Athanasia Kolovou, and Alexandros Potamianos.
\newblock Ntua-slp at semeval-2018 task 2: Predicting emojis using rnns with
  context-aware attention.
\newblock In {\em SemEval@NAACL-HLT}, 2018.

\bibitem{Cappallo2018TheNM}
Spencer Cappallo, Stacey Svetlichnaya, Pierre Garrigues, Thomas Mensink, and
  Cees Snoek.
\newblock The new modality: Emoji challenges in prediction, anticipation, and
  retrieval.
\newblock {\em CoRR}, abs/1801.10253, 2018.

\bibitem{Devlin2018BERTPO}
Jacob Devlin, Ming-Wei Chang, Kenton Lee, and Kristina Toutanova.
\newblock Bert: Pre-training of deep bidirectional transformers for language
  understanding.
\newblock {\em CoRR}, abs/1810.04805, 2018.

\bibitem{Radford2018ImprovingLU}
Alec Radford.
\newblock Improving language understanding by generative pre-training.
\newblock 2018.

\bibitem{Vaswani2017AttentionIA}
Ashish Vaswani, Noam Shazeer, Niki Parmar, Jakob Uszkoreit, Llion Jones,
  Aidan~N. Gomez, Lukasz Kaiser, and Illia Polosukhin.
\newblock Attention is all you need.
\newblock In {\em NIPS}, 2017.

\bibitem{ltekin2018TbingenOsloAS}
Çagri Ç{\"o}ltekin and Taraka Rama.
\newblock T{\"u}bingen-oslo at semeval-2018 task 2: Svms perform better than
  rnns in emoji prediction.
\newblock In {\em SemEval@NAACL-HLT}, 2018.

\end{thebibliography}


\end{document}